\documentclass[letterpaper, 10 pt, conference]{ieeeconf}  % Comment this line out if you need a4paper

\IEEEoverridecommandlockouts                              % This command is only needed if 
\usepackage{graphics} % for pdf, bitmapped graphics files
\usepackage{booktabs}
\usepackage{graphicx}
\usepackage{multirow}
\usepackage{float}
\usepackage{colortbl}
\usepackage{color}
\usepackage[table]{xcolor} % 引入xcolor包，并使用table选项以支持colortbl功能
\definecolor{mygray}{gray}{.88}
\usepackage{bbding}
\usepackage{caption}
\usepackage{pdfpages}
\usepackage{makecell}
\usepackage{amssymb}
\usepackage[backend=biber,style=numeric,sorting=none]{biblatex}
\usepackage{marvosym} 
\usepackage{etoolbox}
\usepackage{hyperref}
\hypersetup{hypertex=true,
colorlinks=true,
linkcolor=blue,
anchorcolor=blue,
citecolor=black}
\usepackage{amsmath} % assumes amsmath package installed
\definecolor{lightblue}{rgb}{0.8,0.85,1} % 浅蓝色
\definecolor{darkblue}{rgb}{0.414902, 0.561765, 0.982353} 
\title{\LARGE \bf
TriHelper: Zero-Shot Object Navigation with Dynamic Assistance
}

\author{Lingfeng Zhang$^{1}$, Qiang Zhang$^{1}$, Hao Wang$^{1}$, Erjia Xiao$^{1}$, Zixuan Jiang$^{1}$, Honglei Chen$^{1}$, Renjing Xu$^{1\dag}$% <-this % stops a space
\thanks{$^{1}$All authors are with The Hong Kong University of Science and Technology (Guangzhou). $^\dag$ is the corresponding author
        {\tt\small renjingxu@hkust-gz.edu.cn}}%
}
\AtEveryBibitem{\small}
\DeclareFieldFormat{labelnumberwidth}{\footnotesize\mkbibbrackets{#1}}

\defbibenvironment{bibliography}
  {\list
     {\printfield[labelnumberwidth]{labelnumber}}
     {% 设置列表参数
      \setlength{\labelwidth}{\labelnumberwidth}%
      \setlength{\leftmargin}{\labelwidth}%
      \setlength{\labelsep}{\biblabelsep}%
      \addtolength{\leftmargin}{\labelsep}%
      \setlength{\itemsep}{\bibitemsep}%
      \setlength{\parsep}{\bibparsep}}}
  {\endlist}
  {\item}
\begin{document}

\maketitle
\thispagestyle{empty}
\pagestyle{empty}

%%%%%%%%%%%%%%%%%%%%%%%%%%%%%%%%%%%%%%%%%%%%%%%%%%%%%%%%%%%%%%%%%%%%%%%%%%%%%%%%
\begin{abstract}

Navigating toward specific objects in unknown environments without additional training, known as Zero-Shot object navigation, poses a significant challenge in the field of robotics, which demands high levels of auxiliary information and strategic planning. Traditional works have focused on holistic solutions, overlooking the specific challenges agents encounter during navigation such as collision, low exploration efficiency, and misidentification of targets. To address these challenges, our work proposes TriHelper, a novel framework designed to assist agents dynamically through three primary navigation challenges: collision, exploration, and detection. Specifically, our framework consists of three innovative components: (i) Collision Helper, (ii) Exploration Helper, and (iii) Detection Helper. These components work collaboratively to solve these challenges throughout the navigation process. Experiments on the Habitat-Matterport 3D (HM3D) and Gibson datasets demonstrate that TriHelper significantly outperforms all existing baseline methods in Zero-Shot object navigation, showcasing superior success rates and exploration efficiency. Our ablation studies further underscore the effectiveness of each helper in addressing their respective challenges, notably enhancing the agent's navigation capabilities. By proposing TriHelper, we offer a fresh perspective on advancing the object navigation task, paving the way for future research in the domain of Embodied AI and visual-based navigation.

\end{abstract}

%%%%%%%%%%%%%%%%%%%%%%%%%%%%%%%%%%%%%%%%%%%%%%%%%%%%%%%%%%%%%%%%%%%%%%%%%%%%%%%%
\section{INTRODUCTION}

Navigating in unknown environments to find a specified target object is a significant challenge in Embodied AI research. 
% The Habitat Challenge, through its indoor Object Goal Navigation task(ObjectNav) benchmark, aims to assess the ability of agents to find specific objects (e.g., bed, TV monitor) within multistorey 3D indoor scenes\cite{savva2019habitat}. In this task, agents must navigate using only information captured by an RGBD camera and global pose data.
The Habitat platform\cite{savva2019habitat}, developed by Facebook AI Research (FAIR), provides a sophisticated simulation environment for this purpose, enabling the testing of AI agents in complex 3D indoor scenes at speeds surpassing real-time. This makes Habitat an ideal platform for tasks like the indoor Object Goal Navigation (ObjectNav) benchmark, which assesses the ability of agents to locate specific objects (e.g., bed, TV monitor) within multi-storey 3D indoor scenes using only RGBD camera captures and global pose data.
  In recent years, to advance ObjectNav, researchers have developed a plethora of 3D scene datasets and navigation methodologies, including reinforcement learning\cite{mazoure2022improving,filos2021psiphi,deitke2022️}, imitation learning\cite{mandi2022towards,Ramrakhya_2023_CVPR}, Zero-Shot learning\cite{yu2023l3mvn,yokoyama2023vlfm,yamauchi1997frontier}, and Few-Shot learning\cite{kok2022few}, each with its specific focus and limitations. In end-to-end training, especially with reinforcement learning, researchers have designed various rewards and penalties for training on scene datasets.

Imitation learning significantly improves task success by teaching agents to navigate in the manner humans search for objects, though it requires extensive training and human demonstrations of the training dataset. Zero-Shot and Few-Shot methods, which represent training with a very small portion of the dataset or without any dataset at all, offer noticeable advantages in deployability and adaptability to different scenes, despite a slight decrease in accuracy compared to fully trained models\cite{yokoyama2023vlfm}.
\begin{figure}[h]
\centering
\includegraphics[height=3.125cm, width=0.475\textwidth]{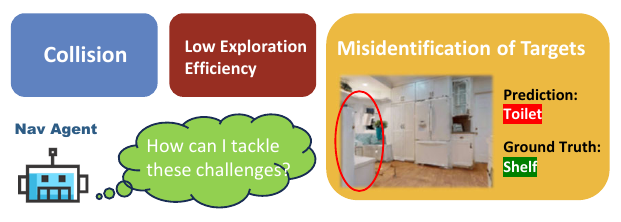}
\caption{Challenges in Zero-Shot Object Navigation.}
% \vspace{-1.01cm}
\label{challenge}
\end{figure}
However, the application of Zero-Shot learning in ObjectNav has encountered significant challenges. Agents often struggle with efficiently navigating towards the target object without prior exposure to similar environments, leading to issues such as frequent collisions, inefficient exploration paths, and inaccurate target identification. 
  
In this study, we conduct a comprehensive analysis of the primary difficulties faced by agents in Zero-Shot navigation tasks, leading to the development of an integrative framework designed to optimize navigation strategies through specific auxiliary modules, thereby enhancing the efficiency of object navigation in unknown multistorey 3D indoor environments. We recognize that while Zero-Shot learning offers the potential for effective navigation without extensive training data, agents still encounter issues such as collision, low exploration efficiency, and misidentification of targets in practical applications, which are shown in \textbf{Fig. \ref{challenge}}. To overcome these challenges, our research not only focuses on a singular navigation strategy but also takes a multi-level dynamic approach, considering the variety of situations an agent might encounter in different settings and designing specialized solutions for each specific problem. Our contributions are specific enhancements for each failure case,  achieving state-of-the-art (SOTA) performance in Zero-Shot learning for the ObjectNav on the Habitat platform. 
  
  Our contributions are summarized as follows:
\begin{itemize}
	\item \textbf{Collision Helper:} We design a clustering algorithm-based auxiliary system after analyzing the agent's motion trajectories and obstacles encountered to address collision issues. 

	\item \textbf{Exploration Helper:} We propose a learning mechanism based on the exploration behavior of agents to tackle the problem of low exploration efficiency. 

	\item \textbf{Detection Helper:} We propose a method employing vision-language models (VLMs) to assist in the detection of target objects from the RGB image, enhancing the accuracy of target detection.

\end{itemize}

Through these innovative approaches, our method not only significantly improves the overall Zero-Shot learning performance of ObjectNav but also provides new perspectives and frameworks for future research, aiming to better address the challenges of visual-based navigation.

\section{RELATED WORK}

\subsection{Object Navigation}

Visual navigation is a critical task for robots, especially in unknown environments. Object Goal Navigation (ObjectNav) focuses on visual navigation within these unknown settings, leveraging semantic priors to enhance a robot's ability to locate objects\cite{batra2020objectnav}. Implementations of the ObjectNav often rely on reinforcement learning\cite{mazoure2022improving,filos2021psiphi,deitke2022️}, imitation learning\cite{cai2023bridging}, or top-down map predictions\cite{chaplot2020object,zhang2021hierarchical,luo2022stubborn}. However, these methods are predominantly based on closed dataset research, making them less applicable to different datasets and platforms. To address the challenges of applying the ObjectNav to various datasets and reduce training consumption, recent developments in Zero-Shot ObjectNav frameworks have garnered significant attention. We will also employ Zero-Shot approaches to conduct ObjectNav.

\subsection{Zero-Shot Object Navigation}

In direct supervision methods, the necessity of retraining due to variations in datasets when transitioning between them significantly increases the training consumption. Zero-Shot ObjectNav effectively addresses this issue and has seen substantial advancements in recent years. For instance, SemExp\cite{chaplot2020object} constructs the semantic map to explore and navigate. CLIP on Wheels (CoW)\cite{gadre2023cows}, by integrating CLIP\cite{radford2021learning} technology, combines egocentric RGB-D images with verbal instructions. 
% It employs zero-shot object localization and a top-down map-updating strategy for exploring environments and precisely locating target objects. 
LGX\cite{dorbala2023can}, LFG\cite{shah2023navigation}, and ESC\cite{zhou2023esc} leverage the common-sense reasoning capabilities of Large Language Models (LLMs) to support sequential navigation decisions. The PixNav\cite{cai2023bridging} project, through the use of foundational models and specifying navigation targets in pixel units, achieves universal navigation for various object types.

However, these methods typically require the conversion of environmental visual information into textual information for target navigation. The application of Visual Language Models (VLMs) offers a novel solution to this issue. VLMs can directly extract semantic information from RGB images and be deployed on consumer-grade laptops, optimizing the processing workflow\cite{yokoyama2023vlfm}. Furthermore, combining VLMs with LLMs for knowledge transfer and the utilization of prior information can significantly enhance the accuracy of locating targets. By integrating the capabilities of VLMs and LLMs—that is, using VLMs to directly extract rich semantic prior information from RGB images while employing LLMs to select the optimal navigation goal points we can effectively improve the success rate of navigation. Compared to existing technologies, this integrative approach demonstrates significant advantages in enhancing navigation efficiency and accuracy.

\section{PRELIMINARY}

\subsection{Problem definition} 

In the framework of the ObjectNav, the agent is tasked with navigating towards 6 specified target object categories, like a 'chair' or 'bed' ($T_i$). The agent starts off at a randomly selected spot within the scenario ($S_i$), with the category of the object ($T_i$) as its target. At every discrete interval, denoted as time step $t$, the agent is equipped with visual data ($V_t$) and readings of its sensor's pose ($P_t$), guiding its choice of movement commands ($a_t$). The visual data encompasses imagery from the agent's point of view, including both color (RGB) and depth details. The space of actions ($\mathcal{A}$) available to the agent includes: {\textit{move\_forward, turn\_left, turn\_right, look\_up, look\_down, and stop}}. Only the \textit{move\_forward} action causes the agent to move 25 cm forward, the other actions only cause the agent to rotate 30 degrees in the corresponding direction, except stop. The agent can opt to stop once it assesses that it has neared the target of interest. Success in an episode is achieved if the agent elects to stop within a predetermined area close to the target object, specifically a distance less than $d_{success}$ (= 0.1 meters). Each episode has a preset limit of 500 time steps, marking the end of an episode.

\subsection{Navigation Map Construction}
\subsubsection{Semantic Map}Semantic map $\mathcal{M}$ is created using RGB-D images and the pose of agents, which is similar to the method in \cite{chaplot2020object}. Its structure is a three-dimensional tensor of size $ C \times W \times H$, where $W \times H$ defines the width and length of the semantic map, and $C$ is equal to $C_n + 5$ where $n$ is the number of object categories, representing the number of channels. In this work, The first four channels of the semantic map respectively represent the obstacle map, explored area, current agent location, and past agent location. The next $n$ channels are the semantic maps with $n$ types of objects $O_{1...n}$. And we finally added a false target semantic map to save misidentified targets for later use. At the time of each episode update, we clear the semantic map, substitute and default the starting position of the agent to the center point of the semantic map $(\frac{W}{2},\frac{H}{2})$. The semantic map is populated by converting visual data into point clouds using geometric techniques, which are then mapped onto a 2D top view. It is characterized by the inclusion of physical obstacles and areas that have been explored, as well as elements of different object species identified by semantic segmentation. The semantic mask matches the point cloud one by one, allowing precise channel mapping on the semantic map. Semantic maps provide the basis for our Zero-Shot ObjectNav.

\begin{figure*}[!htbp]
% \vspace{-1cm}
\centering
\includegraphics[width=1.0\linewidth]{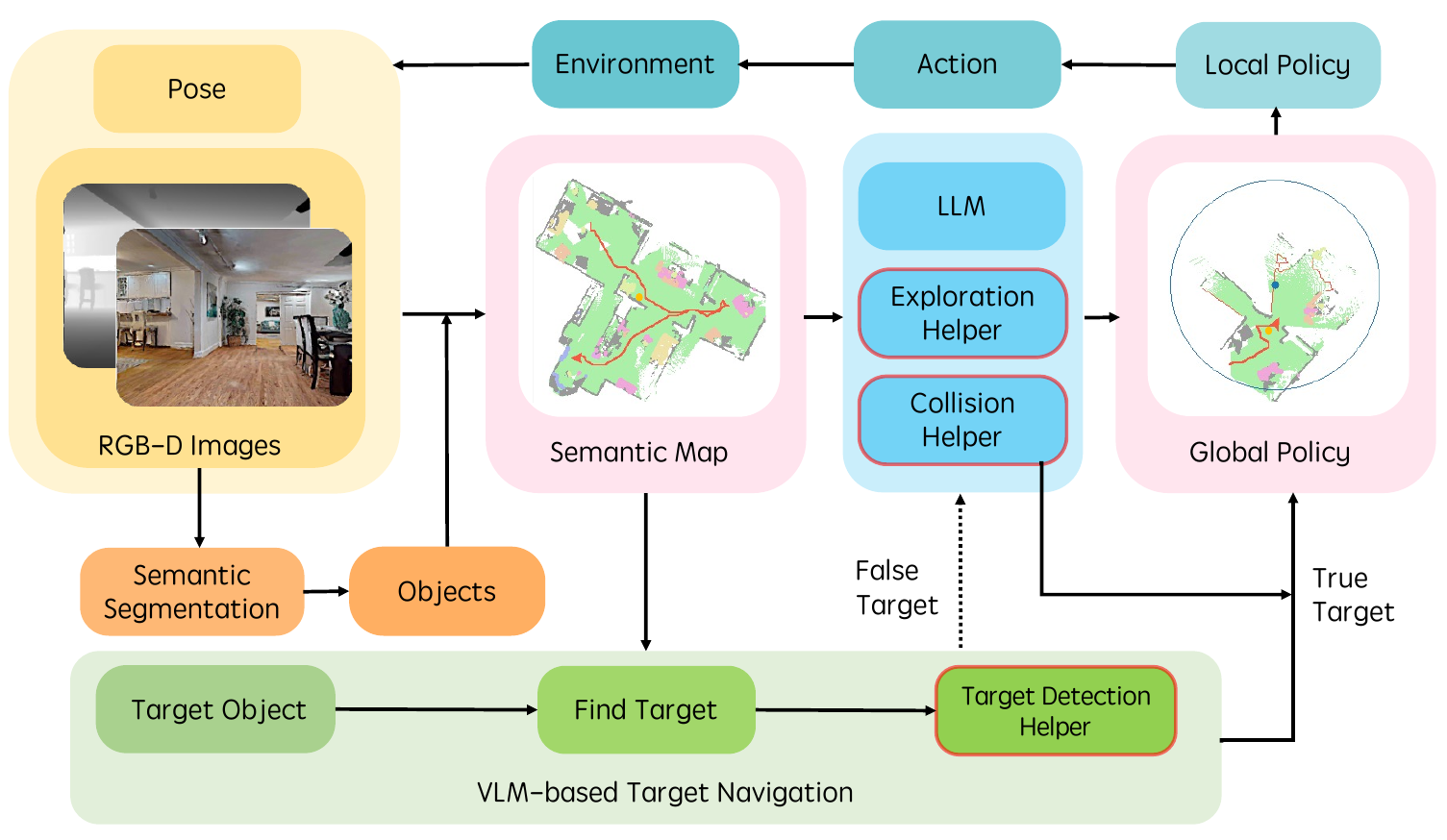}
\caption{The architecture of our framework. At each time step, we first input the RGB-D images into the semantic segmentation module to get the object masks, then construct the semantic map and dynamically use the global policy proposed to select a long-term goal. Finally, we use the local policy to get the action of the agent and interact with the environment. The long-term goal point and the center of the largest connected area of the explorable region are marked in the figure. The dotted line represents the re-entry of the exploration process when the target object is false.}
\vspace{-0.5cm}
\label{framework}
\end{figure*}

\subsubsection{Frontiers Map}
\label{sec:frontier}
After building the semantic map, in order to find the next goal point, we also build the frontier map from the first two channels of the semantic map, using the procedure described in \cite{ramakrishnan2022poni}. The process begins with delineating the boundaries of the explored map to find the outer edges. Expand the edges of the obstacle map, then create the frontier map by subtracting the obstacle map from the explored area. This map highlights potential areas to explore next. Small areas on the map are identified and removed, leaving only significant areas as potential exploration targets. Thus all the remaining centers of the area can be calculated as our frontiers $\mathcal{F}$. A scoring system based on cost and utility proposed in \cite{julia2012comparison} helps prioritize these frontiers, for each frontier cell $F_i \in \mathcal{F}$, we can calculate the score $S^{CU}(F_i)$:
% \vspace{-0.2cm}
\begin{equation}
S^{CU}(F_i) = U(F_i) - \lambda_{CU}C(F_i)
\end{equation}
where $U(F_i)$ is a utility function, $C(F_i)$ is a cost function and the constant $\lambda_{CU}$ adjusts the relative importance between both factors. Each potential frontier cell in these areas is evaluated to determine its viability as an exploration destination, balancing the cost of reaching the destination with the expected utility.

\subsubsection{Frontiers Select}

We utilize the method described in \cite{yu2023l3mvn} to score and select from all the frontiers, which builds a query prompt for each frontier point and utilize an LLM to query:

\textit{``What is the probability that $O_1, O_2,\cdots,O_j, T_i$ exist in a frontier area at the same time?"}

Where $O_{1...j}$ represents all objects contained in the frontiers to which $F_i$ belongs and $T_i$ represents the target object. So we can get the $P(F_{i})$ from the LLM response, and we can find the $F_{LLM}$ such that the function $P(F_{i})$ reaches its maximum value:
\vspace{-0.2cm}
\begin{equation}
F_{{LLM}} = \underset{F_{i}}{\mathrm{argmax}} \ P(F_{i})
\end{equation}
\vspace{-0.1cm}
% After each P, We select the maximum value as the next frontier target point given by LLM as the candidate long-term goal point for global policy selection.
Once $F_{{LLM}}$ is calculated, we utilize it dynamically later in the global policy.

\section{METHODS}
\vspace{-0.15cm}
\subsection{Overview}
The overview of our proposed framework is illustrated in \textbf{Fig. \ref{framework}}. 
Upon acquiring RGB-D images and the agent's pose from the environment, we input the RGB-D images into the semantic segmentation module, obtaining masks for each object category within the images. These masks, along with the original data, are then fed into the mapping module for semantic map construction. After constructing the semantic map, we utilize a global policy mentioned in \ref{sec:global policy} to select the next long-term goal point. After that, we use the local policy for short-term goal navigation mentioned in \ref{sec:policy}. At each time step, we compute a new action for the agent to take and interact with the environment, collecting data for the next time step. Our long-term goal points are updated every 10-time steps to ensure that the agent has enough time to move toward the long-term goal. The VLM-based Target Navigation module is activated only when the target object is detected by the semantic segmentation module, to specifically navigate to the final target object unless the target is false.

\subsection{TriHelper}

\subsubsection{Collision Helper}
We define two types of situations that the agent encounters during the ObjectNav as collisions: First, the agent is trapped in a certain position and collides with nearby objects or obstacles several time steps; second, after setting the long-term goal, the local policy planner mentioned in \ref{sec:policy} cannot compute the reachable path, which hinders the agent from further exploration. To solve the collision problem, we propose a cluster-based collision helper.
 Upon examination, it has been determined that the majority of collision instances stem from the imprudent selection of long-term goals. This often leads to the agent becoming entrapped within confined spaces or failing to identify a navigable path. Therefore, we employ the clustering algorithm to compute the maximum connected space of the navigable area and select the centroid of the maximum connected space as the long-term goal when a collision is detected, so that the agent can reach as broad an area as possible and continue exploring after untrapping.
We use the second channel $M_2$ of the semantic map $\mathcal{M}$ as the navigable area, and the clustering algorithm is as follows:
\begin{equation}
G_{{max}} = \underset{G_{k}}{\mathrm{argmax}} \ |G_k| 
\end{equation}

here ${G_k}$ denotes the set of connected regions into which $M_2$ is decomposed, and $|G_k|$ denotes the size of the computed connected region $G_k$.
After calculating the maximum connected region $G_{max}$, we calculate the center of $G_{max}$:
\begin{equation}
\left\{
\begin{aligned}
x_{\text{centroid}} = \frac{1}{|G_{\text{max}}|} \sum_{(x_i, y_i) \in G_{\text{max}}} x_i \\
y_{\text{centroid}} = \frac{1}{|G_{\text{max}}|} \sum_{(x_i, y_i) \in G_{\text{max}}} y_i
\end{aligned}
\right.
\end{equation}

the calculated center-of-mass coordinates can be used in the subsequent global policy to help the agent untrap.

% Note that the equation is centered using a center tab stop. Be sure that the symbols in your equation have been defined before or immediately following the equation. Use Ò(1)Ó, not ÒEq. (1)Ó or Òequation (1)Ó, except at the beginning of a sentence: ÒEquation (1) is . . .Ó
\subsubsection{Exploration Helper}
After analyzing the ObjectNav of the agent, we observed that the baseline method \cite{yu2023l3mvn}, which utilizes an LLM for frontier selection, exhibits a prolonged failure to update the long-term goal. This issue arises because the $F_{LLM}$ didn't update for a long time. Therefore, we calculate the distance between the agent's current location and the long-term goal point:

\begin{equation}
    d = \sqrt{(x_g - x_a)^2 + (y_g - y_a)^2}
\end{equation}

where $(x_a,y_a)$ denotes the agent's current location, and $(x_g,y_g)$ denotes the long-term goal point.
$d$ is used in the global policy to decide whether to set the LLM to sleep.
% The LLM is transitioned into a dormant state when the distance falls below a predefined threshold $(d < dormant\_thresold)$. During this dormancy, the agent is permitted to engage in stochastic exploration to broaden the explored area. Subsequent to this phase, the LLM is reactivated to resume its function of selecting forward frontiers to serve as long-term objectives.

\subsubsection{Detection Helper}
Semantic segmentation presents a significant challenge in the ObjectNav, where misclassification of the target object can directly precipitate task failure.
% the detection module incorrectly identifies a cabinet as a toilet, which leads to the final failure of the navigation task. 
Consequently, we integrate a VLM to verify the identified target object. Leveraging its capacity for deeper semantic image analysis through prompt-based guidance, the VLM can execute binary classification of the target object by combining it with other objects within the image. Our prompt is set as:

\textit{``Based on all the objects in this picture and the information of this room, judge whether $T_i$ exists in this picture, and output only yes or no."}

% If VLM judges this target object to be true, we navigate directly to it; otherwise, we mask it as a false target and record it in the map to prevent the agent from being fooled again; however, if the agent still does not find the target at the end of this episode, we make the agent navigate to the fake target at the end of the episode to prevent exploration failure. During navigation to the target object, the collision helper is employed to prevent the agent from getting stuck at the end of the navigation. The strategy involves initially directing the agent towards an open area for a brief period before allowing it to proceed towards the target object.

After that, we can get a response from the VLM and give the auxiliary information to the global policy for navigation.

\subsection{Dynamic Policy}

\subsubsection{Global Policy}
\label{sec:global policy}
Here we introduce our global navigation policy, purposed to find the next long-term goal point in the semantic map to direct the agent's trajectory. We use a dynamic policy to select the long-term goal point. First, we get the best frontier goal point $F_{LLM}$ given by LLM from the baseline policy and use it as one of the candidate goal points. Then we check the agent state at the current time step, if the agent chooses the \textit{move\_forward} action and has moved less than 25 cm, or the local policy described later cannot find a feasible path, the agent is put into collision state, and then we use the collision helper to calculate the $G_{max}$ and the $(x_{centroid},y_{centroid})$ as the long term goal point. If not in collision state, we check if $d$ is less than the  $dormant\_thresold$ and $F_{LLM}$ does not change $(F_{LLM\_t} = F_{LLM\_t-1})$, then the exploration helper is enabled to let the LLM sleep for $sleep\_time$ time steps and the agent is free to explore to increase the exploration space. Finally, if the target object is detected by the semantic segmentation module, we use the detection helper to double check the current RGB image. If the target is true, we directly select the detected target object as the final target point and continue to use the collision helper in this period. If the target is false, we mask it out, record it, and let the agent continue to explore. Finally, if the target object is not found after a certain time step $detection\_threshold$ and a false target is recorded, then the false target is marked as the final goal point.

\begin{figure}[h]
\vspace{-0.34cm}
\centering
\includegraphics[width=0.5\textwidth]{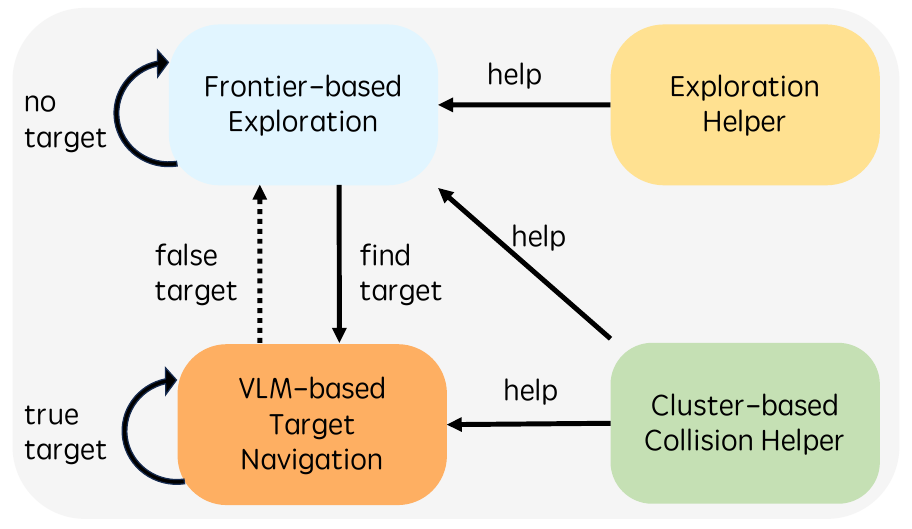}
\caption{The framework of dynamic global policy.}
\vspace{-0.3cm}
\label{modules}
\end{figure}

The three modules we proposed collaborate and complement each other in global policy, and their synergy is illustrated in \textbf{Fig. \ref{modules}}, where they are dynamically utilized throughout the exploration and navigation process to better assist our agent in the ObjectNav.

\subsubsection{Local Policy}
\label{sec:policy}
After finding the long-term goal via the global policy, we deploy the local policy to navigate the agent to the long-term goal point. We use the Fast Marching Method (FMM) \cite{sethian1996fast} for point-to-point navigation configuring the start point at the agent's current locations and the end point at the long-term goal point. FMM calculates the current navigation path at each timestep and then selects the action from the action space based on the agent's current locations, which ensures real-time and effective path planning and navigation. FMM substitutes end-to-end methods such as reinforcement learning, so that our Zero-Shot ObjectNav framework operates without training.

\section{EXPERIMENT}

In this section, we evaluate the performance of our methodology by comparing it to other Zero-Shot ObjectNav baselines. Additionally, We demonstrate the ability of the three agent helpers to solve three failed cases respectively.
\vspace{-0.1cm}
\subsection{Datasets}

Our methodology is evaluated using the Habitat Simulator\cite{savva2019habitat} by applying it to the validation segments of two different datasets: HM3D and Gibson. Specifically, the validation segment for HM3D includes 2000 episodes, spread over 20 scenes. For Gibson, we employ the ObjectNav validation segment developed in SemExp\cite{chaplot2020object} within 5 scenes. The two datasets both encompass 6 target object categories: couch, chair, bed, toilet, TV monitor and potted plant. 
% \vspace{-0.35cm}
\vspace{-0.1cm}
\subsection{Experiment Details}

We evaluate our model on the 3D indoor simulator Habitat platform\cite{savva2019habitat} with an RGB-D image size of (480,640) and an environmental input that also includes the base odometer sensor, a target object represents as an integer of $0-5$. Our implementation is based on Open-Set code in \cite{yu2023l3mvn}. The LLM\cite{liu2019roberta} used in selecting frontiers is calibrated in alignment with the baseline framework. At the same time, we use RedNet\cite{jiang2018rednet} and yolov8\cite{yolov8_ultralytics} to conduct semantic segmentation to predict all existing objects in the RGB-D image. 
For the VLM used in detection helper, we choose the quantized Qwen-VL-Chat-Int4\cite{bai2023qwen} model to ensure effectiveness while improving our inference speed and reducing memory footprint.
For other constants, we set the $dormant\_thresold = 50 cm$, $sleep\_time = 20$ and $detection\_thresold = 400 $.
% \begin{table}[h]
% \caption{An Example of a Table}
% \label{table_example}
% \begin{center}
% \begin{tabular}{|c||c|}
% \hline
% One & Two\\
% \hline
% Three & Four\\
% \hline
% \end{tabular}
% \end{center}
% \end{table}

%    \begin{figure}[thpb]
%       \centering
%       \framebox{\parbox{3in}{We suggest that you use a text box to insert a graphic (which is ideally a 300 dpi TIFF or EPS file, with all fonts embedded) because, in an document, this method is somewhat more stable than directly inserting a picture.
% }}
%       %\includegraphics[scale=1.0]{figurefile}
%       \caption{Inductance of oscillation winding on amorphous
%        magnetic core versus DC bias magnetic field}
%       \label{figurelabel}
%    \end{figure}

% \vspace{-0.35cm}
\vspace{-0.1cm}
\subsection{Evaluation Metrics}

We evaluate our method using Success Rate(SR), Success weighted by Path
Length (SPL) and Distance to Goal (DTG) based on previous work \cite{anderson2018evaluation}. SR is defined as:
$SR = \frac{1}{N} \sum_{i=1}^{N} S_i$, SPL is defined as: $SPL = \frac{1}{N} \sum_{i=1}^{N} S_i \frac{l_i}{\max(p_i, l_i)}$, and DTG is defined as:$DTG = max(||x_T-G||_2 - d_s, 0)$.
Here $N$ represents the total number of evaluated tasks. $S_i$ denotes a binary success indicator for the task, with 1 indicating success (i.e., the agent reached the goal) and 0 indicating failure. $l_i$ is the shortest path length from the starting point to the goal for the $i^{th}$ episode.
$p_i$ is the actual path length traveled by the agent to complete the $i^{th}$ episode. $||x_T-G||_2$ is the L2 distance of the agent from the target object location at the end of the episode, $d_s$ is the success threshold boundary.
The SPL metric, therefore, not only considers whether the agent successfully completes the task but also evaluates the efficiency of the path taken by the agent. The efficiency is gauged by comparing the actual path length ($p_i$) against the shortest possible path length ($l_i$). Thus, higher values are preferable for SR and SPL. DTG represents the distance between the agent and the target objects at the end of episodes, so a lower value is preferable.
\vspace{-0.2cm}
\subsection{Baselines}
To evaluate the Zero-Shot ObjectNav performance of our model, we compare it to several baselines containing the state-of-the-art (SOTA) baseline.

\begin{itemize}

\item \textbf{Randomly Walking:} The agent selects an action randomly from all available actions space $\mathcal{A}$. 

\item \textbf{FBE\cite{yamauchi1997frontier}:} This baseline utilizes a classical robotics framework for constructing maps and selecting frontiers.

\item \textbf{SemExp\cite{chaplot2020object}:} This baseline constructs the semantic map and utilizes reinforcement learning.

\item \textbf{ZSON\cite{majumdar2022zson}:} This work transfers the agent trained in the Image Navigation (ImageNav) task to the ObjectNav. 

\item \textbf{CoW\cite{gadre2023cows}:} CoW applies CLIP\cite{li2022grounded} to detect objects and let the agent always explore the closest frontiers until it finds the target object.  

\item \textbf{ESC\cite{zhou2023esc}:} This work uses the semantic map to navigate and CLIP to detect objects, and implemented LLM to select frontiers. 

\item \textbf{L3MVN\cite{yu2023l3mvn}:} We follow this work as the baseline to construct the semantic map and utilize LLM to select frontiers.

\item \textbf{PixNav\cite{cai2023bridging}:} This baseline uses pixels as navigation targets, trains models for pixel goal navigation, and uses LLMs for long-term navigation planning.

\item \textbf{PONI\cite{ramakrishnan2022poni}:} This baseline determines the location of an unseen target by predicting two latent functions.
\item \textbf{Stubborn\cite{luo2022stubborn}:} This work proposes methods to address collision and detection.
\item \textbf{VLFM\cite{yokoyama2023vlfm}:} In this latest work, researchers built a semantic value map to evaluate frontiers to select exploration directions.

\end{itemize}

\vspace{-0.3cm}
\subsection{Results}
\begin{figure}[!htbp]
\vspace{-0.2cm}
\centering
\includegraphics[width=0.45\textwidth]{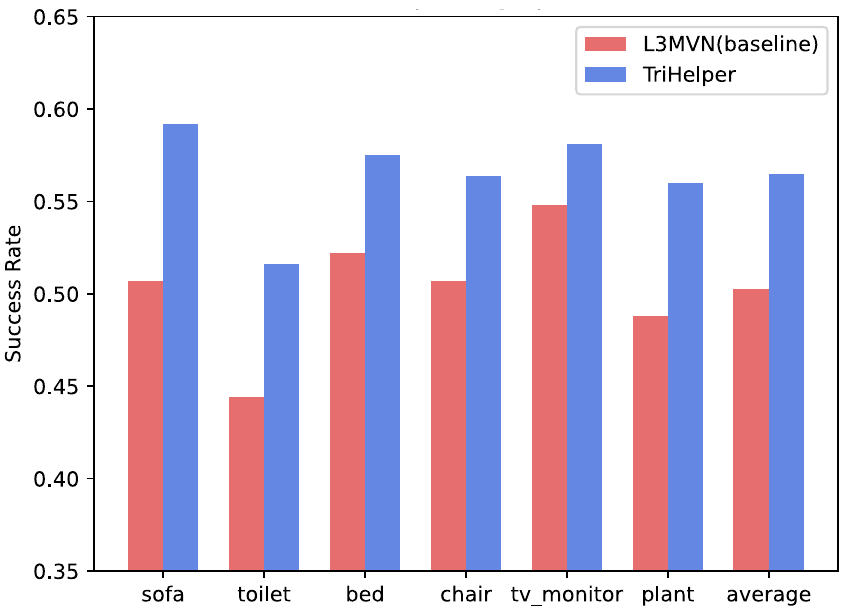}
\caption{The Success Rate by Category and Method.}
\vspace{-0.24cm}
\label{SR}
\end{figure}

The performance of our model on the two datasets HM3D and Gibson is shown in Table \ref{comp}, with empty data indicating that the work was not experimented on that dataset or did not provide the appropriate metrics. TriHelper clearly stands out from all the Zero-Shot ObjectNav methods and achieves SOTA on both datasets. Compared to the baseline we used, our proposed method achieves +6\% SR improvement and +9.5\% SPL on the HM3D dataset; +9.1\% SR improvement and +9.6\% SPL improvement on the Gibson dataset. Compared with the current SOTA model, we achieve +4\% SR improvement on the HM3D dataset and +1.2\% SR improvement on the Gibson dataset. The SR of our method compared to the baseline by object category is illustrated in \textbf{Fig. \ref{SR}}.
The reason why our SPL is slightly lower than that of the SOTA method is that the proposed dynamic global policy focuses on improving the most important metric SR, which indicates the percentage of agents that can successfully navigate to the target object. Since we prevent the agent from choosing a false target for a certain time step, the path length is increased. In addition, since Gibson's scene dataset is more spacious and the task dataset does not contain episodes on the upper and lower floors, our effectiveness in solving the collision problem is diminished, and a small improvement is achieved. We also found that some detection fail cases were caused by simulator misjudgments, so we also carried out a manual double-check on the HM3D, we can see that after removing the misjudgments, our method achieved 5.9\% improvement of baseline and reaches 60\%+ SR. See \ref{appendix} Appendix for detailed analysis.
%In the ablation study, we analyze in detail the proportion of problems solved by each helper to highlight the effectiveness of our approach for Zero-Shot ObjectNav.
\renewcommand{\arraystretch}{1.0}
\begin{table}[ht]
\vspace{-0.2cm}
\scriptsize
\caption{Results of comparative experiment. Our model reaches the SOTA across all the Zero-Shot ObjectNav baselines. We use two color scales of blue to denote the \textbf{\textcolor{darkblue}{first}} and \textbf{\textcolor{lightblue}{second}} best performance. The \textsuperscript{*} denotes that we manually double-check.}
    \centering
    \setlength{\tabcolsep}{1.4mm}{
    \begin{tabular}{cccccccc}
    \toprule[1.0pt]
    
        \multirow{2}{*}{\textbf{Method}} & \multirow{2}{*}{\textbf{Zero-Shot}} & \multicolumn{3}{c}{\textbf{HM3D}}  & \multicolumn{3}{c}{\textbf{Gibson}} \\ \cmidrule(r){3-5} \cmidrule(l){6-8} 
        ~ & ~ & SR$\uparrow$ & SPL$\uparrow$ &DTG$\downarrow$ & SR$\uparrow$ & SPL$\uparrow$ & DTG$\downarrow$ \\ \midrule[1.0pt]
        Randomly & \Checkmark  & 0.000 & 0.000 & 7.600 & 0.030 & 0.030 & 2.580\\ 
        FBE\cite{yamauchi1997frontier} & \Checkmark  & 0.237 & 0.123 & 5.414 & 0.417 & 0.214 & 2.634 \\ 
        SemExp\cite{chaplot2020object} & \XSolidBrush & 0.379 & 0.188 & \cellcolor{darkblue}\textbf{2.943} & 0.652 & 0.336 & 1.520 \\ 
        ZSON\cite{majumdar2022zson} & \XSolidBrush & 0.255 & 0.126 & - & 0.313 & 0.120 & - \\ 
        CoW\cite{gadre2023cows} & \Checkmark  & 0.320 & 0.181 & - & - & - & -\\ 
        ESC\cite{zhou2023esc} & \Checkmark  & 0.385 & 0.220 & - & - & - & - \\ 
        PONI\cite{ramakrishnan2022poni} & \XSolidBrush & - & - & - & 0.736 & 0.410 & 1.250 \\
        
        VLFM\cite{yokoyama2023vlfm} & \Checkmark & \cellcolor{lightblue}{0.525} & \cellcolor{darkblue}\textbf{0.304} & - & \cellcolor{lightblue}{0.840} & \cellcolor{darkblue}\textbf{0.522} & - \\ 
        PixNav\cite{cai2023bridging} & \XSolidBrush & 0.379 & 0.205 & - & - & - & - \\ 
        
        Stubborn\cite{luo2022stubborn} & \XSolidBrush & - & - & - & 0.237 & 0.098 & - \\
        L3MVN\cite{yu2023l3mvn}& \Checkmark  & 0.504 & 0.231 & 4.427 & 0.761 & 0.377 & \cellcolor{lightblue}{1.101} \\
        L3MVN\cite{yu2023l3mvn}\textsuperscript{*}& \Checkmark  & 0.561\textsuperscript{*} & - & - & - & - & - \\ \midrule[1.0pt]
        \rowcolor{mygray}\textbf{TriHelper (Ours)} & \Checkmark & \cellcolor{darkblue}\textbf{0.565} & \cellcolor{lightblue}{0.253} & \cellcolor{lightblue}{3.873} & \cellcolor{darkblue}\textbf{0.852} & \cellcolor{lightblue}{0.431} & \cellcolor{darkblue}\textbf{0.600} \\ 
        \rowcolor{mygray}\textbf{TriHelper (Ours)\textsuperscript{*}} & \Checkmark & \textbf{0.620\textsuperscript{*}} & - & - & - & - & - \\

    \bottomrule[1.0pt]
    \end{tabular}}
    \label{comp}
\vspace{-0.35cm}
\end{table}

% \vspace{-0.15cm}
\subsection{Ablation study}

% \multirow{2}{*}[-0.5ex]{\makecell{\textbf{Solution}\\ \textbf{Percentage (\%)}}}
% {
% \renewcommand{\arraystretch}{1.0}
% \begin{table}[ht]
% \scriptsize
% \caption{Results of comparative experiment. Our model reaches SOTA of all the Zero-Shot ObjectNav baselines.}
%     \centering
%     \setlength{\tabcolsep}{1.4mm}{
%     \begin{tabular}{ccccccc}
%     \toprule[1.0pt]
    
%         \multicolumn{3}{c}{\textbf{TriHelper Ablation}} & \multicolumn{3}{c}{\textbf{Cases Solved}}& \textbf{HM3D Results}  \\ 
%         \cmidrule(r){1-3} \cmidrule(lr){4-6}  \cmidrule(r){7-7} 
%         Helper\_1& Helper\_2& Helper\_3 & Collision& Exploration & Detection & SR\uparrow \\ \midrule[1.0pt]
%         \Checkmark &           &           &   0.030 & 0.535 & 0.000 & 0.000   \\
%          &     \Checkmark      &           &   0.030 & 0.516 & 0.000 & 0.000  \\
%          &           &       \Checkmark    &   0.030 & 0.548 & 0.000 & 0.000  \\
%         \Checkmark &   \Checkmark        &     \Checkmark     &  - & 0.565 & 0.000 & 0.000  \\
        
%     \bottomrule[1.0pt]
%     \end{tabular}}
% \vspace{-0.3cm}
% \end{table}
% }

In order to evaluate the capability of the three proposed helpers for the navigation problem. We conduct ablation experiments on the HM3D dataset: the percentage of corresponding failure cases with different helpers, as well as the improvement in SR. The results in Table \ref{ablation} show that all three of our proposed helpers have significant effects on the corresponding failure cases: The collision helper reduces 6\% of collisions; the exploration helper reduces 2.60\% of failed exploration; and the detection helper reduces 3.05\% of false detection. The best performance is reached when the three helpers are used together: 11.05\% reduction in collision, 2.60\% reduction in false detection, and 6\% improvement in SR compared to baseline. Table \ref{ablation} shows that employing all three helpers simultaneously reduces most collision and many recognition failures, yet exploration failures significantly increase. This issue arises primarily from episodes where the target object is located on a different floor than the agent's initial position, making it challenging for the agent to transition between floors after exploring the initial one. Additionally, the detection helper's masking of false targets can sometimes prevent the agent from successfully locating the target object before the episode ends, contributing to task failure.

% {
% \renewcommand{\arraystretch}{1.0}
% \begin{table}[ht]
% \scriptsize
% \caption{Results of ablation study. Negative numbers denote the increased ratios of corresponding fail cases.}
%     \centering
%     \setlength{\tabcolsep}{1.4mm}{
%     \begin{tabular}{ccccccc}
%     \toprule[1.0pt]
    
%         \multicolumn{2}{c}{\textbf{Collision}} & \multicolumn{2}{c}{\textbf{Exploration}}& \multicolumn{2}{c}{\textbf{Detection}} & \textbf{HM3D Results}  \\ 
%         \cmidrule(r){1-2} \cmidrule(lr){3-4}  \cmidrule(lr){5-6} \cmidrule(r){7-7} 
%         Helper& Solved(\%) & Helper & Solved(\%)& Helper & Solved(\%) & SR\uparrow \\ \midrule[1.0pt]
%         - &       0   &    -       &  0  & - & 0 & 0.505  \\ 
%         \Checkmark &       31.41    &    -       &  -21.98  & - & -2.06 & 0.530   \\
%          -&      -    &    \Checkmark       &  -  &-  & - & 0  \\
%          -&     -4.17      &   -     &   11.72 & \Checkmark & \textbf{17.99}& 0.540  \\
%         \Checkmark &   \textbf{57.85}        &     \Checkmark    &  -55.68 & \Checkmark & 15.63 & \textbf{0.565}  \\
        
%     \bottomrule[1.0pt]
%     \end{tabular}}
% \vspace{-0.5cm}
% \end{table}
% }
{
\renewcommand{\arraystretch}{1.0}
\vspace{-0.4cm}
\begin{table}[ht]
\scriptsize
\caption{Results of ablation study on HM3D. The $\downarrow and \uparrow$ respectively denote the decrease and increase in the number of corresponding failures. The \textsuperscript{*} denotes that we manually double-check.}
    \centering
    \resizebox{\linewidth}{!}{
    \begin{tabular}{ccccc}
    \toprule[1.0pt]
    
        \multirow{2}{*}{\textbf{Ablation}} & \textbf{Collision}& \textbf{Exploration}& \textbf{Detection} & \textbf{HM3D Results}  \\ 
        % \cmidrule(r){2-2} \cmidrule(lr){3-4}  
         &\textbf{Failure(\%)}  & \textbf{Failure(\%)}& \textbf{Failure(\%)} & \textbf{SR$\uparrow$} \\ \midrule[1.0pt]
        No Helper &       19.10        &  13.65   & 16.95 & 0.505  \\ \midrule
        Collision &      13.10$\downarrow $         &  16.65$\downarrow$   & 17.30$\uparrow$ & 0.530   \\
        Exploration&      18.45$\downarrow$     &  \textbf{11.05}$\downarrow$   & 17.35$\uparrow$ & 0.531  \\
         Detection&     20.00$\uparrow$       &   12.05$\downarrow$  & \textbf{13.90}$\downarrow$& 0.540  \\ \midrule
        \rowcolor{mygray}\textbf{TriHelper} &   \textbf{8.05}$\downarrow $       &  21.25$\uparrow$  & 14.30$\downarrow $& \textbf{0.565}  \\
         \rowcolor{mygray}\textbf{TriHelper}\textsuperscript{*} &   \textbf{8.05}$\downarrow$       &  21.25$\uparrow $ & \textbf{8.80}\textsuperscript{*}$\downarrow$ & \textbf{0.620}\textsuperscript{*} \\
    \bottomrule[1.0pt]
    \end{tabular}}
\label{ablation}
\vspace{-0.4cm}
\end{table}
}
% \vspace{-0.05cm}
\section{CONCLUSIONS}

In this work, we presented TriHelper, a novel framework aimed at addressing the key challenges of Zero-Shot ObjectNav. Our framework substantially enhances the navigational proficiency of autonomous agents operating within unfamiliar environments by dynamically incorporating a trio of synergistic components: Collision Helper, Exploration Helper, and Detection Helper. Each component is meticulously designed to tackle specific problems encountered during navigation, namely collision avoidance, efficient exploration, and accurate target detection.

Our extensive experiments conducted on the HM3D and Gibson datasets have demonstrated the superior performance of TriHelper against all the existing baseline methods. The ablation studies further validated the critical role of each helper component, highlighting their collective importance in enhancing the agent's navigation success.

By integrating these innovative solutions, TriHelper offers a robust and dynamic framework that significantly advances the field of Zero-Shot ObjectNav. Our approach underscores the necessity of targeted assistance and strategic planning in overcoming navigation challenges, highlighting the importance of targeted assistance for overcoming specific navigation challenges. Further research could focus on solving the exploration problem caused by exploring different floors.


%%%%%%%%%%%%%%%%%%%%%%%%%%%%%%%%%%%%%%%%%%%%%%%%%%%%%%%%%%%%%%%%%%%%%%%%%%%%%%%%

%%%%%%%%%%%%%%%%%%%%%%%%%%%%%%%%%%%%%%%%%%%%%%%%%%%%%%%%%%%%%%%%%%%%%%%%%%%%%%%%

%%%%%%%%%%%%%%%%%%%%%%%%%%%%%%%%%%%%%%%%%%%%%%%%%%%%%%%%%%%%%%%%%%%%%%%%%%%%%%%%
% \clearpage

\printbibliography

@inproceedings{yu2023l3mvn,
  title={L3mvn: Leveraging large language models for visual target navigation},
  author={Yu, Bangguo and Kasaei, Hamidreza and Cao, Ming},
  booktitle={2023 IEEE/RSJ International Conference on Intelligent Robots and Systems (IROS)},
  pages={3554--3560},
  year={2023},
  organization={IEEE}
}

@article{sethian1996fast,
  title={A fast marching level set method for monotonically advancing fronts.},
  author={Sethian, James A},
  journal={proceedings of the National Academy of Sciences},
  volume={93},
  number={4},
  pages={1591--1595},
  year={1996},
  publisher={National Acad Sciences}
}

@inproceedings{savva2019habitat,
  title={Habitat: A platform for embodied ai research},
  author={Savva, Manolis and Kadian, Abhishek and Maksymets, Oleksandr and Zhao, Yili and Wijmans, Erik and Jain, Bhavana and Straub, Julian and Liu, Jia and Koltun, Vladlen and Malik, Jitendra and others},
  booktitle={Proceedings of the IEEE/CVF international conference on computer vision},
  pages={9339--9347},
  year={2019}
}

@article{bai2023qwen,
  title={Qwen-vl: A versatile vision-language model for understanding, localization, text reading, and beyond},
  author={Bai, Jinze and Bai, Shuai and Yang, Shusheng and Wang, Shijie and Tan, Sinan and Wang, Peng and Lin, Junyang and Zhou, Chang and Zhou, Jingren},
  year={2023}
}

@article{anderson2018evaluation,
  title={On evaluation of embodied navigation agents},
  author={Anderson, Peter and Chang, Angel and Chaplot, Devendra Singh and Dosovitskiy, Alexey and Gupta, Saurabh and Koltun, Vladlen and Kosecka, Jana and Malik, Jitendra and Mottaghi, Roozbeh and Savva, Manolis and others},
  journal={arXiv preprint arXiv:1807.06757},
  year={2018}
}

@inproceedings{luo2022stubborn,
  title={Stubborn: A strong baseline for indoor object navigation},
  author={Luo, Haokuan and Yue, Albert and Hong, Zhang-Wei and Agrawal, Pulkit},
  booktitle={2022 IEEE/RSJ International Conference on Intelligent Robots and Systems (IROS)},
  pages={3287--3293},
  year={2022},
  organization={IEEE}
}

@inproceedings{li2022grounded,
  title={Grounded language-image pre-training},
  author={Li, Liunian Harold and Zhang, Pengchuan and Zhang, Haotian and Yang, Jianwei and Li, Chunyuan and Zhong, Yiwu and Wang, Lijuan and Yuan, Lu and Zhang, Lei and Hwang, Jenq-Neng and others},
  booktitle={Proceedings of the IEEE/CVF Conference on Computer Vision and Pattern Recognition},
  pages={10965--10975},
  year={2022}
}

@article{liu2019roberta,
  title={Roberta: A robustly optimized bert pretraining approach},
  author={Liu, Yinhan and Ott, Myle and Goyal, Naman and Du, Jingfei and Joshi, Mandar and Chen, Danqi and Levy, Omer and Lewis, Mike and Zettlemoyer, Luke and Stoyanov, Veselin},
  journal={arXiv preprint arXiv:1907.11692},
  year={2019}
}

@software{yolov8_ultralytics,
  author = {Glenn Jocher and Ayush Chaurasia and Jing Qiu},
  title = {Ultralytics YOLOv8},
  version = {8.0.0},
  year = {2023},
  url = {https://github.com/ultralytics/ultralytics},
  orcid = {0000-0001-5950-6979, 0000-0002-7603-6750, 0000-0003-3783-7069},
  license = {AGPL-3.0}
}

@inproceedings{yamauchi1997frontier,
  title={A frontier-based approach for autonomous exploration},
  author={Yamauchi, Brian},
  booktitle={Proceedings 1997 IEEE International Symposium on Computational Intelligence in Robotics and Automation CIRA'97.'Towards New Computational Principles for Robotics and Automation'},
  pages={146--151},
  year={1997},
  organization={IEEE}
}

@article{jiang2018rednet,
  title={Rednet: Residual encoder-decoder network for indoor rgb-d semantic segmentation},
  author={Jiang, Jindong and Zheng, Lunan and Luo, Fei and Zhang, Zhijun},
  journal={arXiv preprint arXiv:1806.01054},
  year={2018}
}

@article{chaplot2020object,
  title={Object goal navigation using goal-oriented semantic exploration},
  author={Chaplot, Devendra Singh and Gandhi, Dhiraj Prakashchand and Gupta, Abhinav and Salakhutdinov, Russ R},
  journal={Advances in Neural Information Processing Systems},
  volume={33},
  pages={4247--4258},
  year={2020}
}

@article{majumdar2022zson,
  title={Zson: Zero-shot object-goal navigation using multimodal goal embeddings},
  author={Majumdar, Arjun and Aggarwal, Gunjan and Devnani, Bhavika and Hoffman, Judy and Batra, Dhruv},
  journal={Advances in Neural Information Processing Systems},
  volume={35},
  pages={32340--32352},
  year={2022}
}

@article{julia2012comparison,
  title={A comparison of path planning strategies for autonomous exploration and mapping of unknown environments},
  author={Juli{\'a}, Miguel and Gil, Arturo and Reinoso, Oscar},
  journal={Autonomous Robots},
  volume={33},
  pages={427--444},
  year={2012},
  publisher={Springer}
}

@inproceedings{ramakrishnan2022poni,
  title={Poni: Potential functions for objectgoal navigation with interaction-free learning},
  author={Ramakrishnan, Santhosh Kumar and Chaplot, Devendra Singh and Al-Halah, Ziad and Malik, Jitendra and Grauman, Kristen},
  booktitle={Proceedings of the IEEE/CVF Conference on Computer Vision and Pattern Recognition},
  pages={18890--18900},
  year={2022}
}

@inproceedings{gadre2023cows,
  title={Cows on pasture: Baselines and benchmarks for language-driven zero-shot object navigation},
  author={Gadre, Samir Yitzhak and Wortsman, Mitchell and Ilharco, Gabriel and Schmidt, Ludwig and Song, Shuran},
  booktitle={Proceedings of the IEEE/CVF Conference on Computer Vision and Pattern Recognition},
  pages={23171--23181},
  year={2023}
}

@inproceedings{zhou2023esc,
  title={Esc: Exploration with soft commonsense constraints for zero-shot object navigation},
  author={Zhou, Kaiwen and Zheng, Kaizhi and Pryor, Connor and Shen, Yilin and Jin, Hongxia and Getoor, Lise and Wang, Xin Eric},
  booktitle={International Conference on Machine Learning},
  pages={42829--42842},
  year={2023},
  organization={PMLR}
}

@article{cai2023bridging,
  title={Bridging zero-shot object navigation and foundation models through pixel-guided navigation skill},
  author={Cai, Wenzhe and Huang, Siyuan and Cheng, Guangran and Long, Yuxing and Gao, Peng and Sun, Changyin and Dong, Hao},
  journal={arXiv preprint arXiv:2309.10309},
  year={2023}
}

@inproceedings{yokoyama2023vlfm,
  title={Vlfm: Vision-language frontier maps for zero-shot semantic navigation},
  author={Yokoyama, Naoki Harrison and Ha, Sehoon and Batra, Dhruv and Wang, Jiuguang and Bucher, Bernadette},
  booktitle={2nd Workshop on Language and Robot Learning: Language as Grounding},
  year={2023}
}

@article{mazoure2022improving,
  title={Improving zero-shot generalization in offline reinforcement learning using generalized similarity functions},
  author={Mazoure, Bogdan and Kostrikov, Ilya and Nachum, Ofir and Tompson, Jonathan J},
  journal={Advances in Neural Information Processing Systems},
  volume={35},
  pages={25088--25101},
  year={2022}
}

@inproceedings{filos2021psiphi,
  title={Psiphi-learning: Reinforcement learning with demonstrations using successor features and inverse temporal difference learning},
  author={Filos, Angelos and Lyle, Clare and Gal, Yarin and Levine, Sergey and Jaques, Natasha and Farquhar, Gregory},
  booktitle={International Conference on Machine Learning},
  pages={3305--3317},
  year={2021},
  organization={PMLR}
}

@inproceedings{mandi2022towards,
  title={Towards more generalizable one-shot visual imitation learning},
  author={Mandi, Zhao and Liu, Fangchen and Lee, Kimin and Abbeel, Pieter},
  booktitle={2022 International Conference on Robotics and Automation (ICRA)},
  pages={2434--2444},
  year={2022},
  organization={IEEE}
}

@article{kok2022few,
  title={A few-shot learning-based reward estimation for mapless navigation of mobile robots using a siamese convolutional neural network},
  author={Kok, Vernon and Olusanya, Micheal and Ezugwu, Absalom},
  journal={Applied Sciences},
  volume={12},
  number={11},
  pages={5323},
  year={2022},
  publisher={MDPI}
}

@article{batra2020objectnav,
  title={Objectnav revisited: On evaluation of embodied agents navigating to objects},
  author={Batra, Dhruv and Gokaslan, Aaron and Kembhavi, Aniruddha and Maksymets, Oleksandr and Mottaghi, Roozbeh and Savva, Manolis and Toshev, Alexander and Wijmans, Erik},
  journal={arXiv preprint arXiv:2006.13171},
  year={2020}
}

@inproceedings{zhang2021hierarchical,
  title={Hierarchical object-to-zone graph for object navigation},
  author={Zhang, Sixian and Song, Xinhang and Bai, Yubing and Li, Weijie and Chu, Yakui and Jiang, Shuqiang},
  booktitle={Proceedings of the IEEE/CVF international conference on computer vision},
  pages={15130--15140},
  year={2021}
}

@inproceedings{radford2021learning,
  title={Learning transferable visual models from natural language supervision},
  author={Radford, Alec and Kim, Jong Wook and Hallacy, Chris and Ramesh, Aditya and Goh, Gabriel and Agarwal, Sandhini and Sastry, Girish and Askell, Amanda and Mishkin, Pamela and Clark, Jack and others},
  booktitle={International conference on machine learning},
  pages={8748--8763},
  year={2021},
  organization={PMLR}
}

@article{dorbala2023can,
  title={Can an Embodied Agent Find Your “Cat-shaped Mug”? LLM-Based Zero-Shot Object Navigation},
  author={Dorbala, Vishnu Sashank and Mullen Jr, James F and Manocha, Dinesh},
  journal={IEEE Robotics and Automation Letters},
  year={2023},
  publisher={IEEE}
}

@inproceedings{shah2023navigation,
  title={Navigation with large language models: Semantic guesswork as a heuristic for planning},
  author={Shah, Dhruv and Equi, Michael Robert and Osi{\'n}ski, B{\l}a{\.z}ej and Xia, Fei and Ichter, Brian and Levine, Sergey},
  booktitle={Conference on Robot Learning},
  pages={2683--2699},
  year={2023},
  organization={PMLR}
}

@article{deitke2022️,
  title={ProcTHOR: Large-Scale Embodied AI Using Procedural Generation},
  author={Deitke, Matt and VanderBilt, Eli and Herrasti, Alvaro and Weihs, Luca and Ehsani, Kiana and Salvador, Jordi and Han, Winson and Kolve, Eric and Kembhavi, Aniruddha and Mottaghi, Roozbeh},
  journal={Advances in Neural Information Processing Systems},
  volume={35},
  pages={5982--5994},
  year={2022}
}

@InProceedings{Ramrakhya_2023_CVPR,
    author    = {Ramrakhya, Ram and Batra, Dhruv and Wijmans, Erik and Das, Abhishek},
    title     = {PIRLNav: Pretraining With Imitation and RL Finetuning for ObjectNav},
    booktitle = {Proceedings of the IEEE/CVF Conference on Computer Vision and Pattern Recognition (CVPR)},
    month     = {June},
    year      = {2023},
    pages     = {17896-17906}
}
% \bibliography{main}

\clearpage
\onecolumn 
\section{APPENDIX}
\label{appendix}

\subsection{Manually Double-check}
\begin{figure*}[!ht]
\centering
\includegraphics[width=1.0\linewidth]{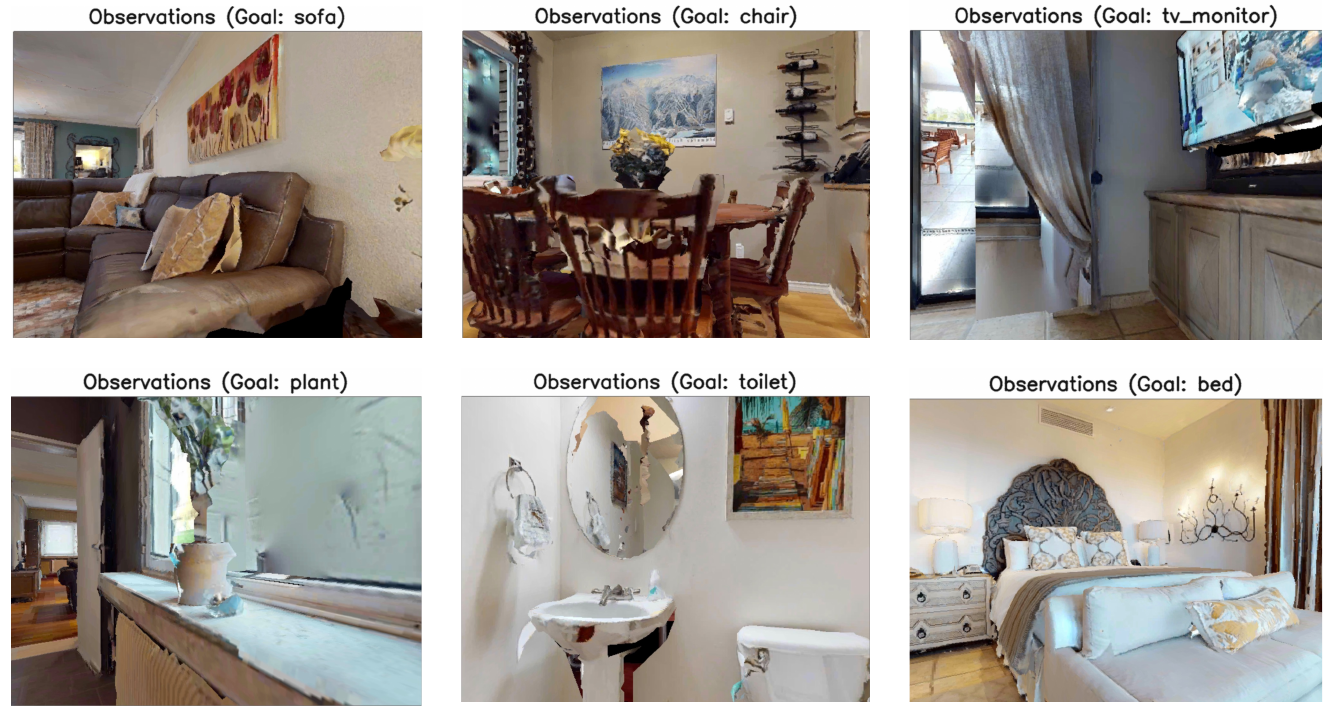}
\caption{The misjudgments of the simulator.}
\vspace{-0.3cm}
\label{appendix1}
\end{figure*}

After an in-depth analysis of the results of each episode, we found that some of the failures due to detection were caused by misjudgment of the simulator, as shown in \textbf{Fig. \ref{appendix1}}. In order to correct these misjudgments, we manually double-check the HM3D experiment results of the TriHelper and baseline for detection cases. The results are shown in the Table \ref{apendix} below.
{
\renewcommand{\arraystretch}{1.0}
\begin{table}[ht]
\scriptsize
\caption{Results of manual double-check on HM3D.}
    \centering
    \scalebox{1.5}{
    \begin{tabular}{cccc}
    \toprule[1.0pt]
    
        \textbf{Method} & \textbf{Detection Cases/ Corrected Cases}& \textbf{Detection Ratio Corrected(\%)}& \textbf{HM3D SR}  \\ 
        \cmidrule(r){1-1} \cmidrule(lr){2-2} \cmidrule(r){3-3} \cmidrule(r){4-4} 
        % Helper& Solved(\%) & Helper & Solved(\%)& Helper & Solved(\%) & SR\uparrow \\ \midrule[1.0pt]
        L3MVN\cite{yu2023l3mvn} &      339/112     &  33.04   &  0.561\\
        \textbf{TriHelper (Ours)}&      286/110    &   38.41  &    \textbf{0.620} \\

    \bottomrule[1.0pt]
    \end{tabular}}
\vspace{0.3cm}
\label{apendix}
\end{table}
}

Through the table data, we can analyze that there are 33.04\%, 38.41\% of simulator misjudgment cases respectively in the baseline and our model, and after subtracting these cases, we achieve 0.561, 0.620 of SR respectively. At this time, our model improves 5.91\% of SR compared to the baseline. We can see that after adding the detection helper, our model is able to solve the problem of target object misrecognition very well.

\end{document}